\documentclass[runningheads]{llncs}
\usepackage{amsmath, amssymb, amsfonts, booktabs, multirow}
\usepackage{graphicx,verbatim}
\usepackage{xcolor}

\usepackage{bm}

\newcommand{\EE}{\mathbb{E}}
\newcommand{\RR}{\mathbb{R}}
\newcommand{\uu}{\bm{u}}

\begin{document}
%
%\title{Energy-Based Tissue Manifolds: Personalized multiparametric MRI analysis for tracking tissue-at-risk}

% \title{Tracking tissue changes: Patient-specific multiparametric MRI tissue manifolds for monitoring tissue-at-risk in neurooncology}

\title{Energy-based Tissue Manifolds for Longitudinal Multiparametric MRI Analysis}

\titlerunning{Energy-Based Tissue Manifolds}

% \author{Anonymized Authors}
% \authorrunning{Anonymized Author et al.}
% \institute{Anonymized Affiliations \\
%     \email{email@anonymized.com}}

\author{
Kartikay Tehlan\inst{1,2,3,4}\and
Lukas F\"orner\inst{1,2,3,4} \and
Sina Wendrich\inst{1} \and \\
Nico Schmutzenhofer\inst{1} \and
Michael Fr\"uhwald\inst{5} \and
Matthias Wagner\inst{1,6} \and
Nassir Navab\inst{3} \and
Thomas Wendler\inst{1,2,3,4,6}}
\authorrunning{K. Tehlan et al.}
\institute{Dept. of diagnostic and interventional Radiology and Neuroradiology, University Hospital Augsburg, Germany \\ \textbf{\email{kartikay.tehlan@med.uni-augsburg.de}}\and
Digital Medicine, University Hospital Augsburg, Germany \and
Chair for Computer Aided Medical Procedures and Augmented Reality, Technical University of Munich, Germany
\and
Bavarian Center for Cancer Research (BZKF) Augsburg, Germany \and
Dept. of Pediatrics and Adolescent Medicine, University Hospital Augsburg, Germany \and
Center for Advanced Analytics and Predictive Sciences, University of Augsburg, Germany}

\maketitle
\begin{abstract}
We propose a geometric framework for longitudinal multi-parametric MRI analysis based on patient-specific energy modelling in sequence space. Rather than operating on images with spatial networks, each voxel is represented by its multi-sequence intensity vector ($T1$, $T1c$, $T2$, FLAIR, ADC), and a compact implicit neural representation is trained via denoising score matching to learn an energy function $E_\theta(\uu)$ over $\mathbb{R}^d$ from a single baseline scan.
The learned energy landscape provides a differential-geometric description of tissue regimes without segmentation labels. Local minima define tissue basins, gradient magnitude reflects proximity to regime boundaries, and Laplacian curvature characterises local constraint structure. Importantly, this baseline energy manifold is treated as a fixed geometric reference: it encodes the set of contrast combinations observed at diagnosis and is not retrained at follow-up.
Longitudinal assessment is therefore formulated as evaluation of subsequent scans relative to this baseline geometry. Rather than comparing anatomical segmentations, we analyse how the distribution of MRI sequence vectors evolves under the baseline energy function. In a paediatric case with later recurrence, follow-up scans show progressive deviation in energy and directional displacement in sequence space toward the baseline tumour-associated regime before clear radiological reappearance. In a case with stable disease, voxel distributions remain confined to established low-energy basins without systematic drift.
The presented cases serve as proof-of-concept that patient-specific energy manifolds can function as geometric reference systems for longitudinal mpMRI analysis without explicit segmentation or supervised classification, providing a foundation for further investigation of manifold-based tissue-at-risk tracking in neuro-oncology. The code is available at:  \url{https://github.com/tkartikay/EnFold-MRI/}.

\end{abstract}


%\keywords{Energy-based models \and Multi-parametric MRI \and Longitudinal analysis \and Tissue manifold \and Implicit neural representations \and Pediatric ATRT}

\section{Introduction}

Multi-parametric MRI (mpMRI) is the clinical standard for brain tumour assessment. The joint acquisition of T1, T1c, T2, FLAIR, and diffusion-derived ADC/DWI sequences enables evaluation of tumour microstructure and treatment response. Computational approaches typically treat these sequences as multi-channel images for segmentation or as inputs to supervised classification models. Both paradigms operate in anatomical space and depend on labelled data or predefined class structures, assumptions that are difficult to sustain in longitudinal settings where disease evolution is gradual and patient-specific.

We consider a complementary perspective. For a fixed acquisition and preprocessing pipeline, each voxel is represented by a sequence vector
\[
\uu = (u_{\mathrm{T1}}, u_{\mathrm{T1c}}, u_{\mathrm{T2}}, u_{\mathrm{FLAIR}}, u_{\mathrm{ADC}})^\top \in \mathbb{R}^d,
\]
which we interpret as a point in \emph{sequence space}. Across a patient's brain, these vectors do not occupy $\mathbb{R}^d$ arbitrarily but concentrate on a structured subset determined by tissue composition, partial volume effects, and measurement physics, with distinct tissue states corresponding to localised regions within this space.

We propose to model this structured subset using a patient-specific energy function $E_\theta : \mathbb{R}^d \rightarrow \mathbb{R}$, learned via denoising score matching on voxel-wise sequence vectors from a single baseline scan. The low-energy region defines the patient's \emph{tissue manifold}: a geometric reference encoding the set of biologically plausible contrast combinations at diagnosis under a given protocol. In this landscape, energy local minima correspond to stable tissue regimes, ridges separate them, and the differential structure of $E_\theta$ characterises transitions between them.

\begin{figure}[h!]
\includegraphics[width=\textwidth]{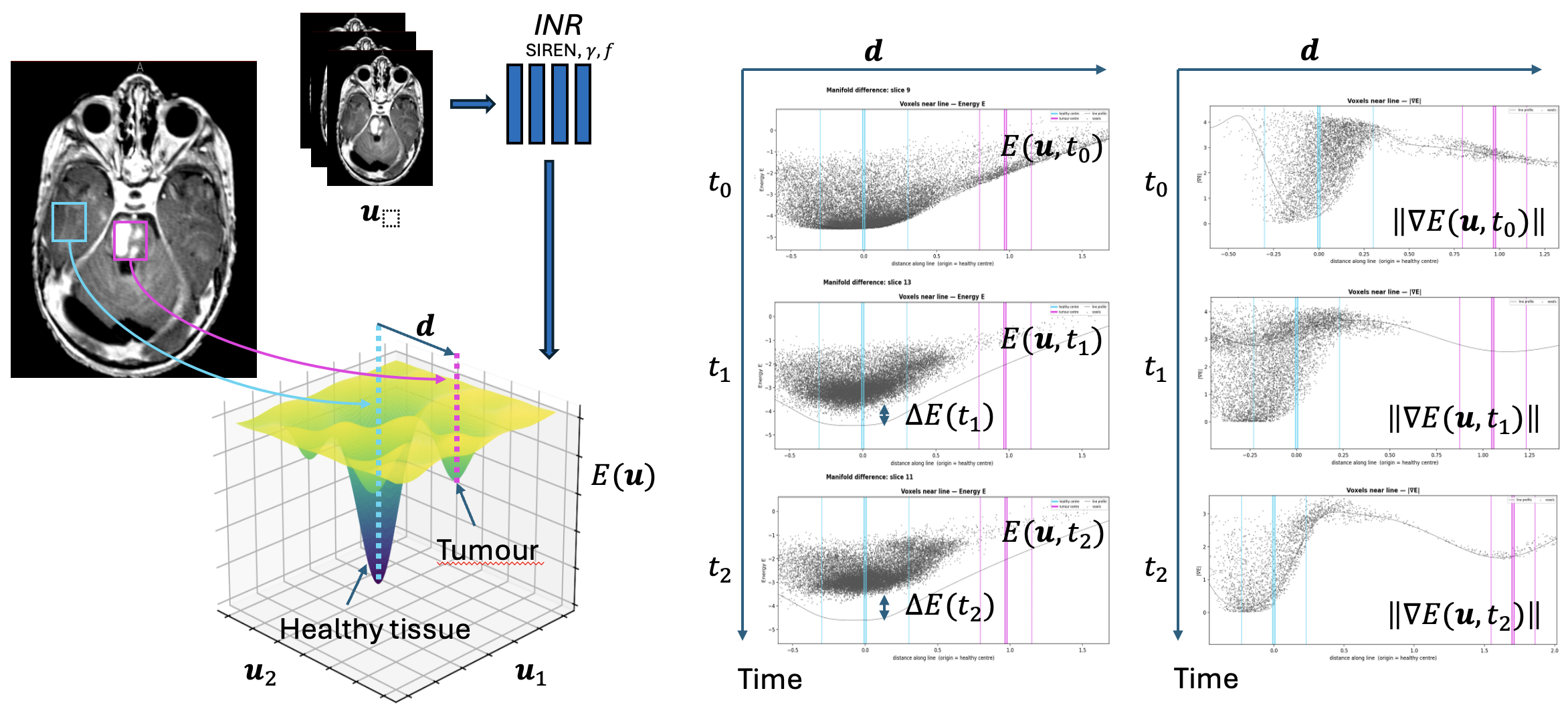}
\caption{Overview of the proposed energy-based longitudinal tissue tracking framework.
\textit{Left:} Two regions of interest (ROIs) are manually placed on the baseline scan ($t_0$): one in healthy tissue (cyan) and one covering the tumour (magenta). Their centroids in 5-dimensional sequence space define two basin attractors of the learned energy function $E_\theta(\uu)$, visualised here in 2D for clarity. The vector $\bm{d}$ connecting the healthy tissue centroid to the tumour centroid defines a one-dimensional axis in sequence space along which longitudinal analysis is performed.
\textit{Centre:} For each time point, brain voxels are projected onto $\bm{d}$ and their energies $E(\uu, t_k)$ are displayed as scatter plots. The shift $\Delta E(t_k)$ relative to the baseline manifold $E(\uu, t_0)$ quantifies how far follow-up tissue states deviate from the original configuration.
\textit{Right:} The corresponding gradient magnitude $\|\nabla E(\uu, t_k)\|$ along $\bm{d}$ reflects changes in basin boundaries and ridge stability over time.
Tracking the distribution of voxel energies and their displacement along the healthy--tumour axis enables detection of tissue at risk.}
\label{fig:graphical_abstract}
\end{figure}

The geometric reference forms the longitudinal hypothesis in this work. We treat the baseline energy landscape as a fixed reference and evaluate follow-up scans relative to it, without retraining. Two quantities are considered. First, for each follow-up scan, we evaluate the distribution of sequence vectors under the baseline energy function. An increase in energy relative to baseline indicates that follow-up tissue states are less compatible with the original configuration, reflecting deviation from the prior tissue intensity distribution without implying pathology by itself. Second, at baseline, healthy and tumour regimes occupy distinct regions in sequence space. After resection, the tumour regime is anatomically absent but remains geometrically defined in the baseline representation. We quantify whether follow-up sequence vectors exhibit displacement in sequence space toward the baseline tumour region, measured purely with respect to the baseline geometry and without assuming spatial correspondence  (Fig. \ref{fig:graphical_abstract}).

%In stable disease, follow-up vectors remain largely confined to existing low-energy basins and show no systematic displacement toward the baseline tumour regime. In recurrence, follow-up vectors show both elevated energy relative to the baseline manifold and increased occupation of regions geometrically proximal to the baseline tumour regime.

%After surgical resection, disease stability or recurrence can be evaluated by analysing how follow-up sequence vectors distribute relative to the baseline energy landscape. In stable disease, vectors remain within existing low-energy basins and do not shift toward the baseline tumour regime. In recurrence, vectors exhibit increased energy relative to the baseline reference and directional drift in sequence space toward the prior tumour region.

%This sequence with longitudinal segmentation pipelines, where early recurrence may be poorly defined and label boundaries are not available or stable. Here, longitudinal assessment is reframed as geometric tracking in sequence space rather than comparison of anatomical masks.

This formulation differs from longitudinal segmentation pipelines, where recurrence assessment relies on evolving anatomical masks whose boundaries may be ill-defined, and label boundaries are neither available nor stable.  Here, longitudinal evaluation is reframed as measurement of geometric deviation in sequence space relative to a fixed patient-specific reference, rather than comparison of anatomical masks.

The contributions of this work are as follows:
(i) we introduce patient-specific energy-defined tissue manifolds in MRI sequence space as a geometric reference representation for longitudinal analysis;
(ii) we show that gradient-flow basins of the learned energy function yield interpretable tissue regimes without segmentation labels, enabling manual anchoring of healthy and tumour regions at baseline;
(iii) in two paediatric brain tumour case studies, we show that longitudinal evolution relative to the baseline manifold distinguishes stable disease from recurrence before radiological evidence of it.

This work is explicitly exploratory. We do not propose a predictive clinical model, but a representation framework for longitudinal tissue analysis in sequence space, opening the way for early detection of tissue at risk in oncology.

\begin{comment}

Our contribution is stated as follows:

(1) We introduce patient-specific energy-defined tissue sets in MRI sequence space as a geometric reference representation.

(2) We show that gradient-flow basins of the learned energy function yield interpretable tissue regimes without segmentation labels, enabling manual anchoring of healthy and tumour regions at baseline.

(3) We demonstrate, in pediatric ATRT-SHH cases, that longitudinal evolution relative to the baseline manifold distinguishes stable disease from recurrence in a proof-of-concept evaluation.

This work is exploratory. We do not propose a predictive clinical model, but a representation framework for longitudinal tissue analysis in sequence space.
\end{comment}

\section{Related Work}
Multi-parametric MRI has traditionally been analysed in anatomical space. Classical approaches model voxel intensities using Gaussian mixture models and hidden Markov random fields, partitioning tissue into predefined classes under spatial smoothness assumptions. More recent methods employ deep convolutional networks for segmentation or classification, learning spatial representations from labelled datasets. These approaches treat multi-sequence inputs as multi-channel images and rely either on parametric assumptions or supervised training \cite{deverdier20242024braintumorsegmentation,kazerooni2024braintumorsegmentationpediatrics}.

In parallel, generative modelling has advanced substantially in medical imaging. Denoising diffusion probabilistic models (DDPMs) and score-based generative models learn data distributions through noise-perturbed training objectives, recovering images via iterative reverse processes \cite{ho2020denoising,yang2023diffusion}. Energy-based models (EBMs) instead represent data distributions through a scalar energy function that assigns low energy to observed samples and higher energy elsewhere, enabling flexible modelling without explicit normalization \cite{du2019implicit}. In medical imaging, diffusion and score-based methods have primarily been applied in image space for generation, reconstruction, and inverse problems, integrating learned priors with acquisition physics \cite{yang2023diffusion}.

Our work differs in two respects. First, we operate in sequence space rather than image space: the object of interest is the distribution of voxel-wise contrast vectors, not spatial image patches. Second, the learned energy function is used as a geometric reference for longitudinal analysis rather than for image synthesis or reconstruction. The goal is not generation, but characterisation of tissue regimes and measurement of geometric deviation relative to a baseline manifold. In this sense, our formulation is related to score-based and energy-based modelling in its training objective, but distinct in its operating space, scale, and longitudinal purpose.

% Denoising Diffusion Prob Models \cite{ho2020denoising}
% Review of diffusion \cite{yang2023diffusion}
% Energy based models \cite{du2019implicit}
% DSM \cite{vincent2011connection}
% how to train with DSM \cite{song2021train}

%============================================================================
\section{Method}
%============================================================================

\subsection{Tissue as Energy in Sequence Space}
Consider a patient with $d$ registered MRI sequences. Each voxel $i$ yields a sequence vector $\uu_i = (u_i^{(1)}, \ldots, u_i^{(d)})^\top \in \RR^d$, where $d=5$ in our setting (T1, T1c, T2, FLAIR, ADC). The collection $\mathcal{U} = \{\uu_i\}_{i=1}^{N}$ of all brain voxels defines an empirical distribution $p_{\text{data}}$ supported on a low-dimensional tissue manifold $\mathcal{M} \subset \RR^d$. We seek to learn an energy function $E_\theta: \RR^d \to \RR$ such that $p_\theta(\uu) \propto \exp(-E_\theta(\uu))$ approximates $p_{\text{data}}$.

\subsection{Score Matching on Sequence Vectors}
Direct maximum likelihood estimation of $E_\theta$ is intractable due to the partition function. We instead train via \emph{denoising score matching} (DSM)~\cite{vincent2011connection}. Let $\bm{a} = \uu + \bm{\epsilon}$, where $\bm{\epsilon} \sim \mathcal{N}(\bm{0}, \sigma^2 \bm{I})$ is isotropic Gaussian noise. The DSM objective is:
\begin{equation}
\mathcal{L}(\theta) = \EE_{\uu \sim p_{\text{data}},\, \bm{\epsilon} \sim \mathcal{N}(\bm{0}, \sigma^2 \bm{I})} \left[ \left\| -\nabla_{\bm{a}} E_\theta(\bm{a}) + \frac{\bm{\epsilon}}{\sigma^2} \right\|^2 \right].
\label{eq:dsm}
\end{equation}
Minimising Eq.~\eqref{eq:dsm} trains the score function $\bm{s}_\theta(\uu) = -\nabla E_\theta(\uu)$ to match the score of the noise-convolved data distribution, without requiring computation of the partition function \cite{song2021train,vincent2011connection}. Unlike image-space applications that require multi-scale noise schedules to handle high dimensionality~\cite{song2019generative}, our low-dimensional sequence space ($d=5$) permits effective training with a single noise level $\sigma$, selected to match the scale of natural intensity variation across tissue types.

\subsection{Architecture: $\gamma$-INR Energy Network}
We parameterise $E_\theta$ as an implicit neural representation with Fourier feature encoding and sinusoidal activations:
\begin{equation}
E_\theta(\uu) = f_L \circ f_{L-1} \circ \cdots \circ f_1 \circ \gamma(\uu),
\end{equation}
where $\gamma(\uu) = [\sin(2\pi \bm{B}\uu),\, \cos(2\pi \bm{B}\uu)]^\top$ is a Gaussian Fourier feature encoding~\cite{tancik2020fourier} with learnable frequencies $\bm{B} \in \RR^{m \times d}$, and each $f_\ell$ is a SIREN layer~\cite{sitzmann2020implicit}: $f_\ell(\bm{x}) = \sin(\omega_0 (\bm{W}_\ell \bm{x} + \bm{b}_\ell))$. The sinusoidal activations ensure that $E_\theta$ and its derivatives, needed for score matching, gradient, and Laplacian computation, are smooth and well-defined everywhere. The 4 hidden layer compact INR with 256 frequencies encoding enables patient-specific training in less than 5 minutes on a MacBook with an M3 Pro (12-core CPU, Metal 3, 36GB RAM).

\subsection{Geometric Characterisation of Tissue Regimes}
After training, $E_\theta(\uu)$ defines a smooth scalar field over sequence space. Tissue regimes correspond to basins of attraction under the gradient flow $\dot{\uu}=-\nabla E_\theta(\uu)$ \cite{milnor}. Local minima define basin attractors, and the energy barriers between them define regime separation. The gradient magnitude and Laplacian provide quantitative descriptors of boundary steepness and basin curvature. This baseline geometry serves as a fixed reference for all longitudinal analyses below.

\section{Case Studies in Paediatric Brain Tumours}
\subsection{Study Design}
We analysed longitudinal mpMRI scans from two paediatric brain tumour patients. For each patient, the diagnostic pre-resection scan ($t_0$) was used to train a patient-specific energy manifold in $\mathbb{R}^5$ sequence space.
To anchor healthy tissue and tumour basins without requiring segmentation masks, we manually placed two regions of interest (ROIs) at $t_0$ for each patient: one covering a representative portion of the tumour, and one in a healthy tissue region. The sequence-space centroids of these ROIs define the reference positions of the tumour and healthy tissue basins, and the line connecting them in sequence space provides a one-dimensional axis along which longitudinal displacement is measured (Fig. \ref{fig:graphical_abstract}).
Subsequent scans ($t_1$--$t_k$) were projected into the original sequence-space energy landscape without retraining, for longitudinal tissue-at-risk tracking.

\subsection{Energy-Based Longitudinal Trajectories}
For each follow-up scan, we evaluated:
(i) the mean energy of healthy-cluster voxels relative to the original manifold;
(ii) basin width and barrier height along the healthy--tumour centre axis in sequence space;
(iii) directional drift in sequence space toward the original tumour basin.

\noindent\textbf{Stable disease:}
In one patient followed for two years after resection, healthy tissue remained within the original low-energy basin. Energy levels and barrier heights remained stable, and no drift toward the tumour regime was observed.

\begin{figure}
    \centering
    \includegraphics[width=0.65\linewidth]{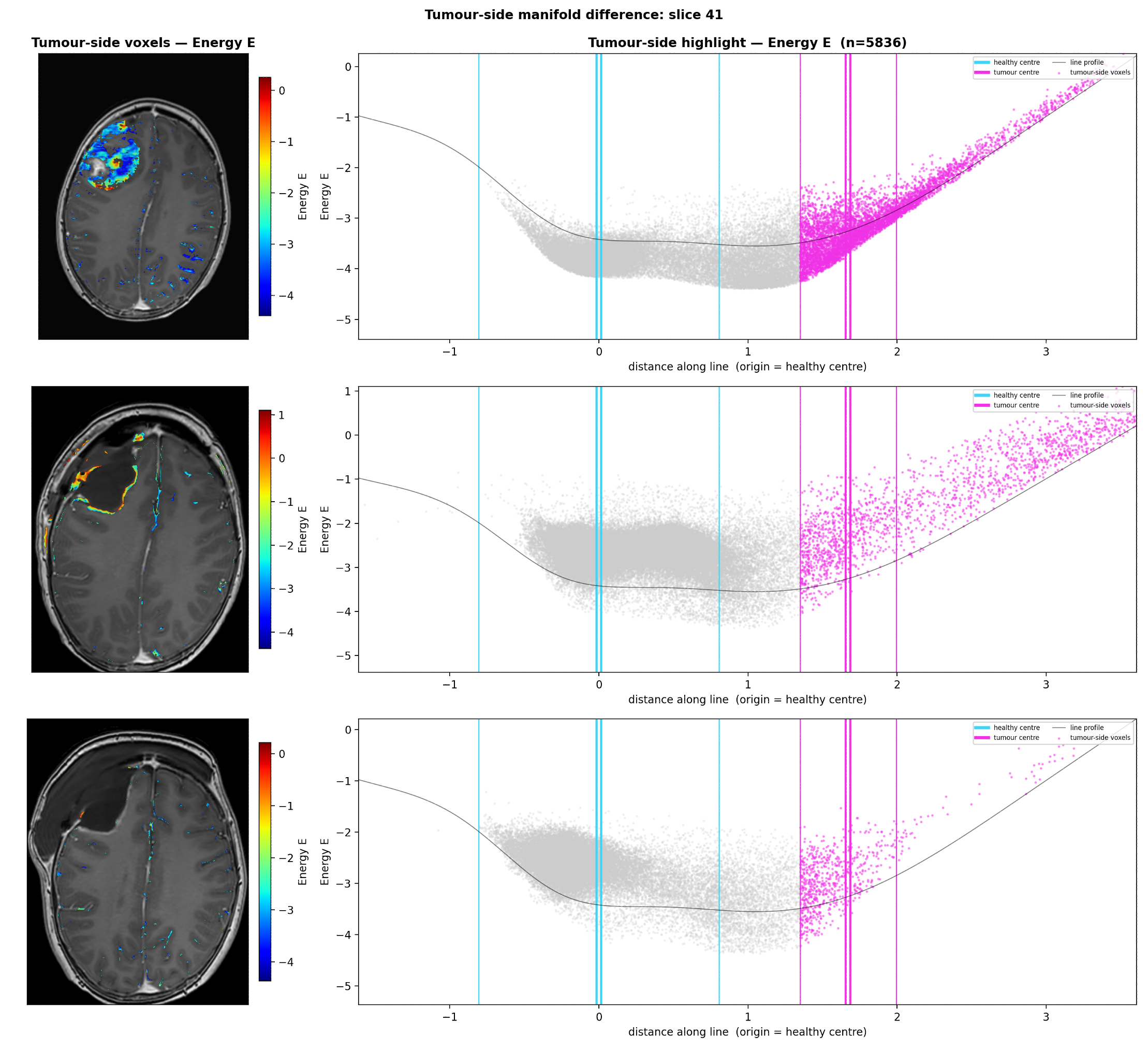}
    \caption{Longitudinal projection of voxel energies along the baseline healthy–tumour axis for a patient with stable disease. The baseline manifold (top) defines the reference geometry; follow-up scans (middle, bottom) show preserved basin structure and stable energy distribution without progressive shift toward the baseline tumour regime, consistent with absence of recurrence.}
    \label{fig:stable}
\end{figure}

\noindent\textbf{Recurrence:}
In a second patient, prior to radiologically evident recurrence, voxels demonstrated progressive displacement in sequence space toward the original tumour basin. Energy relative to the baseline manifold increased, and basin geometry changed, including widening and reduced ridge stability. At the time of confirmed recurrence, a new tumour basin emerged in sequence space. These observations suggest that energy-manifold drift may precede visually segmentable tumour recurrence.

\begin{figure}
    \centering
    \includegraphics[width=0.8\linewidth]{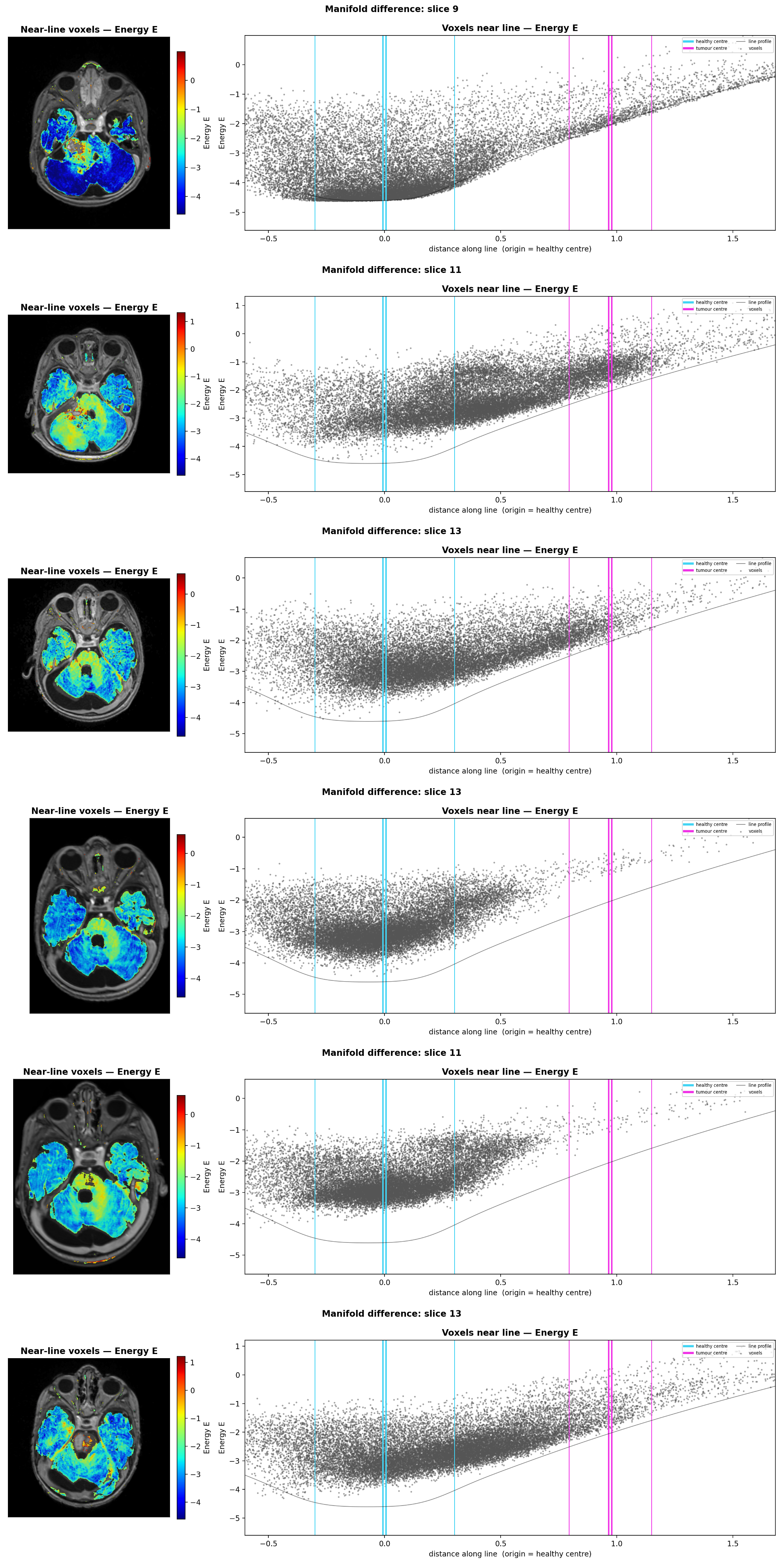}
    \caption{Longitudinal projection of voxel energies along the baseline healthy–tumour axis for a patient with recurrence. Relative to the baseline manifold, follow-up scans show progressive redistribution of voxel energies and displacement toward the baseline tumour regime, accompanied by deformation of the energy profile, consistent with re-emergence of tumour-associated sequence states prior to clear anatomical delineation.}
    \label{fig:recurrence}
\end{figure}

\begin{table}[h]
\centering
%\caption{Differences between each time point and the baseline scan for both patients. $\delta E$ denotes mean energy change relative to baseline; negative values indicate reduced energy compared to baseline. Drift is measured as Euclidean distance in $\mathbb{R}^5$ projected onto the healthy--tumour centroid axis: negative values indicate displacement away from the baseline tumour location in sequence space, positive values indicate displacement toward it. Time points correspond to: immediately after resection (within 48 hours, $t_1$), two months after resection ($t_2$), and the last available diagnostic scan ($t_3$), which is 2 years post-resection for the stable disease patient and 13 months post-resection for the recurrence patient. All differences are statistically significant ($p \ll 0.05$).}

\caption{Mean energy change ($\delta E$) and sequence-space drift relative to baseline for both patients at three time points: immediately post-resection ($t_1$, $<$48h), two months post-resection ($t_2$), and last available scan ($t_3$; 2 years and 13 months post-resection for stable and recurrence patients, respectively). Drift is projected onto the healthy--tumour centroid axis in $\mathbb{R}^5$; negative values indicate displacement away from the tumour basin, positive values toward it. All differences are statistically significant ($p \ll 0.05$).}
\label{tab1}
\begin{tabular}{l|l|l|l|l}
    \hline
    Patient & $\delta E_{t1}$ & $\delta E_{t2}$ & $\delta E_{t3}$ & Drift \\
    \hline
    Stable disease & $0.827$ & $0.684$ & $-0.063$ & $-0.081$ \\
     & $(\pm0.009)$ & $(\pm0.008)$ & $(\pm 0.360)$ & $(\pm 0.005)$ \\
    \hline
    Recurrence & $1.119$ & $+0.880$ & $+0.925$ & $+0.258$ \\
    & $(\pm0.009)$ & $(\pm0.008)$ & $(\pm 0.006)$ & $(\pm 0.004)$ \\
    \hline
\end{tabular}
\end{table}

\section{Discussion}

This work presents a geometric framework for longitudinal interpretation of mpMRI in sequence space. It does not propose a supervised classifier, nor does it aim to demonstrate predictive deployment. Instead, it establishes a patient-specific reference structure in the form of an energy-defined manifold learned from a diagnostic baseline scan. This manifold serves as a geometric representation of the tissue regimes that are compatible with the patient’s baseline mpMRI.

By training an energy function over voxel-wise sequence vectors at baseline, we obtain a smooth scalar field whose minima define tissue regimes as basins of attraction under gradient flow. The geometry of this field provides a structured coordinate system in sequence space. Subsequent scans are evaluated relative to this reference rather than re-segmented independently. In this formulation, longitudinal analysis becomes the study of how new contrast vectors relate to the original energy geometry.

Two forms of longitudinal change can then be examined. First, one may measure energy-relative deviation, that is, whether new voxel vectors exhibit increased energy with respect to the baseline manifold, indicating reduced compatibility with previously observed tissue regimes. Second, one may quantify directional movement in sequence space, in particular, drift toward the region corresponding to the original tumour regime. In the recurrence case analysed, a new basin emerged in sequence space, and voxel vectors demonstrated directional movement toward the prior tumour location defined at baseline. In the stable case, post-operative scans remained confined to existing low-energy basins without systematic drift toward the tumour-associated regime. These observations suggest that recurrence can be interpreted as a geometric reorganisation in sequence space, potentially preceding or accompanying visible anatomical change.

The proposed framework differs from standard clustering approaches in that it does not merely partition points at a given time point. The learned energy function defines a continuous scalar field with differential structure. Basin depth reflects regime stability, barrier height reflects separation between regimes, curvature encodes local constraint structure, and gradient flow describes directional transitions. These geometric quantities provide a basis for longitudinal comparison that is not dependent on segmentation boundaries, which may be unstable or poorly defined in early recurrence.

The present study is limited to a small number of pediatric brain tumor cases and should be regarded as a proof-of-concept methodological investigation rather than a clinical validation. Future work should assess the robustness of this geometric tracking paradigm across larger cohorts, evaluate its sensitivity to early recurrence, and explore integration with spatial information. Population-level comparison of sequence-space energy geometries may further clarify shared and disease-specific structure in tissue organisation over time.

\begin{credits}
\subsubsection{\ackname} This research was partially funded by the Intramural Research Funding Grants ``AI-driven Longitudinal Lesion Tracking'' and ``Precision Medicine for pHGG'' of the Faculty of Medicine, University of Augsburg, German Children Cancer Foundation under grant A 2024/05/DKS 2025.01, the Bavarian Center for Cancer Research as part of the Lighthouse ``Local Therapies'' and the Study Group ``Surrogate parameters for CNS Tumors in Childhood'', as well as by the Bavarian Ministry of Economic Affairs, Regional Development and Energy (StMWi) under grant number DIK-2310-0004//DIK0556/02.
\end{credits}

\begin{comment}
\subsubsection{\discintname}
It is now necessary to declare any competing interests or to specifically
state that the authors have no competing interests. Please place the
statement with a bold run-in heading in small font size beneath the
(optional) acknowledgments\footnote{If EquinOCS, our proceedings submission
system, is used, then the disclaimer can be provided directly in the system.},
for example: The authors have no competing interests to declare that are
relevant to the content of this article. Or: Author A has received research
grants from Company W. Author B has received a speaker honorarium from
Company X and owns stock in Company Y. Author C is a member of committee Z.

\end{comment}

\newpage
\bibliographystyle{splncs04}
\bibliography{refs}

\end{document}